\documentclass[twocolumn,svgnames]{article}

\usepackage{microtype}
\usepackage{xurl} 
\usepackage{natbib}
\usepackage{amsmath,amsthm}
\usepackage{paralist}
\usepackage{listings}
\usepackage{lmodern}
\usepackage{marvosym}
\usepackage[hidelinks,pdfauthor={Giuseppe Macario},pdftitle={An Open, Cross-Platform, Web-Based Metaverse Using WebXR and A-Frame},pdfkeywords={https://orcid.org/0000-0003-4820-155X}]{hyperref}
\usepackage{orcidlink}

\begin{document}
\title{An Open, Cross-Platform, Web-Based Metaverse \\ Using WebXR and A-Frame}
\author{
Giuseppe Macario\,\orcidlink{0000-0003-4820-155X}\,\href{mailto:gm@mit.edu.it}{\textcolor{LightSkyBlue}{\Letter}} \\[2mm]
\small
\begin{tabular}{c|c}
Universitas Mercatorum & Ministry of Defense \\
Piazza Mattei 10, Rome, Italy & Via XX Settembre 8, Rome, Italy
\end{tabular}
}

\maketitle

\begin{abstract}
The metaverse has received much attention in the literature and industry in the last few years, but the lack of an open and cross-platform architecture has led to many distinct metaverses that cannot communicate with each other. This work proposes a WebXR-based cross-platform architecture for developing spatial web apps using the A-Frame and Networked-Aframe frameworks with a view to an open and interoperable metaverse, accessible from both the web and extended reality devices. A prototype was implemented and evaluated, supporting the capability of the technology stack to enable immersive experiences across different platforms and devices. Positive feedback on ease of use of the immersive environment further corroborates the proposed approach, underscoring its effectiveness in facilitating engaging and interactive virtual spaces. By adhering to principles of interoperability and inclusivity, it lives up to Tim Berners-Lee's vision of the World Wide Web as an open platform that transcends geographical and technical boundaries.

\end{abstract}

\section{Introduction}

The advent of extended reality (XR)---namely augmented reality (AR), mixed reality (MR), and virtual reality (AR)---together with a renewed interest in the metaverse after Facebook Inc. changed its name to Meta Platforms Inc., introduces not only unprecedented opportunities for the digital landscape but also significant challenges due to the considerable variation in capabilities between XR devices and conventional computing platforms such as smartphones, tablets, and desktop computers. Where once the primary considerations were screen size and input method (e.g. touch versus mouse and keyboard), developers now must navigate a complex array of sensory inputs, immersive environments and spatial interactions. Therefore, while the web expands to encompass these new immersive experiences, it has become necessary to find the right balance between embracing the unique features of each platform and ensuring universal accessibility. Not only would this allow the web to harness the full potential of XR technologies, but creators could also deliver rich and immersive experiences while preserving the user's freedom to choose their preferred platform and interface.

Tim Berners-Lee, inventor of the World Wide Web, wanted his creature to be an ``open platform that allows anyone to share information, access opportunities and collaborate across geographical boundaries'' \citep{Solon2017FutureWeb}. To achieve this result, \citet{CERN2024HistoryWeb} states that the web should be based on open standards, avoiding proprietary systems. The web indeed owes much of its success to the adoption of standards that allow content to be enjoyed by a wide audience, regardless of the browser or the operating system. Due to this inclusive and flexible nature, the web has already become a major platform for consuming two-di\-men\-sion\-al content.

On the other hand, the journey of the web towards a universally accessible platform has not been without its challenges. Instances where proprietary technologies were widely adopted, included but not limited to the ones developed by Microsoft in the 1990s (e.g. ActiveX), exemplify the potential pitfalls of deviating from open standards. This approach led to a significant portion of web content being optimized exclusively for specific browsers, notably Internet Explorer for Windows, thereby limiting accessibility across different platforms. Similarly, the advent of mobile computing devices, including smartphones and tablets, initially posed significant obstacles for web usability due to the vast differences in screen sizes and interaction models compared to traditional desktop computers. However, the web community's commitment to inclusivity and universal access spurred the evolution of standards to overcome these challenges: for example, CSS emerged as a fundamental solution, enabling web content to dynamically adapt to a highly diverse array of devices.

In the growing area of virtual reality and digital interconnectedness, the vision of an open metaverse---a vast, shared, and interoperable virtual space---has captivated the imagination of technologists, creators, and users alike. This work sets out to propose a cross-platform architecture designed to build such an open metaverse, delivered through the World Wide Web. Central to this exploration is the following research question:
\newtheorem*{rquest}{RQ}
\begin{rquest}
How can a WebXR-based cross-platform architecture facilitate the development of an open, accessible, and interoperable metaverse, and what are the implications for user engagement within this digital ecosystem?
\end{rquest}

The inquiry probes into the viability of leveraging WebXR technology, alongside the integration of the A-Frame and Networked-Aframe frameworks, which enhance the WebXR ecosystem by simplifying the creation of interactive 3D content and enabling real-time, multiplayer experiences across various devices.

\section{Background}

\subsection{Metaverse}
\label{section:metaverse}

When the American writer Neal Stephenson first described a virtual environment named Metaverse in a 1990s sci-fi novel, little did he know that the product of his imagination would soon become reality: Active Worlds (1995) and Second Life (2003) are the first two popular actual forms of Stephenson's Metaverse.\footnote{The Metaverse (with an initial capital letter) is the proper noun of the environment described by Neal Stephenson; in this paper, the metaverse (common noun in lower case) will denote any actual implementation in computer science, distinct from Stephenson's fictional Metaverse.} Since then, the general public has familiarized with the concept of metaverse in information technology, and several computer scientists have given various definitions, for instance \citet{Ritterbusch2023DefiningMetaverse}:
 \newtheorem{definition}{Definition}
\begin{definition}[Metaverse]
A (decentralized) three-di\-men\-sion\-al online environment that is persistent and immersive, in which users represented by avatars can participate socially and economically with each other in a creative and collaborative manner, in virtual spaces decoupled from the real physical world.
\end{definition}

Updating a prior definition of virtual world by \citet{Bell2008VirtualWorlds}, the metaverse can also be defined more concisely as
\begin{definition}[Metaverse]
A persistent synchronous virtual environment, shared by people represented as avatars, facilitated by networked devices.
\end{definition}

These definitions do not explicitly mention XR\slash VR\slash MR\slash AR\footnote{From now on, XR\slash VR\slash MR\slash AR will be abbreviated to XR.} nor do they require interoperability between different metaverses, which makes sense because most implementations, especially the older ones (including the aforementioned Active Worlds and Second Life), do not have XR capabilities and are incompatible with each other. On the other hand, after 30 years since the inception of the metaverse, technology is mature enough to handle these new demands: this is where WebXR and A-Frame come in.

\subsection{WebXR}
\label{section:webxr}

The W3C Immersive Web Working Group has been established to address the challenges of crafting a standard named WebXR Device API; its stated mission is ``to help bring high-performance virtual reality (VR) and augmented reality (AR), collectively known as extended reality (XR), via APIs to interact with XR devices and sensors \emph{in browsers}'' \citep{W3C2024ImmersiveWebCharter}.  As a matter of fact, the centrality of browsers as gateways to digital information has remained a constant since the inception of the World Wide Web: recognizing this key role, the World Wide Web Consortium (W3C) has made the strategic decision to leverage the browser as a foundational platform for spatial web apps (see section \ref{section:spatial}) that render immersive environments. WebXR spatial web apps can run on any device equipped with a modern browser, enabling a wide audience to engage with 3D environments without necessarily resorting to specialized software or hardware.

The WebXR Device API is designed to offer a unified, platform-agnostic abstraction layer that grants access through the web to the fundamental features shared by all XR devices. By abstracting the complexities of underlying hardware, the WebXR Device API enable web developers to access a broad spectrum of interaction controllers through a streamlined interface while simplifying the real-time rendering of immersive environments.

The foundational layer for graphical rendering in WebXR applications is WebGL, i.e. a JavaScript API that leverages the capabilities of OpenGL for Embedded Systems (OpenGL ES) to render interactive 2D and 3D graphics directly within web browsers. WebGL provides a direct access to the GPU through a low-level interface, enabling high-performance rendering of 3D graphics and animations. In order to abstract away the complexities of direct WebGL programming, WebXR applications typically use a higher-level open-source framework or library, to choose from 
\begin{compactdesc}
\item[Babylon.js] Open-source framework (Apache license) by Microsoft and other contributors,
\item[Three.js] Open-source framework (MIT license) by Ricardo Cabello and other contributors,
\item[A-Frame] Open-source framework (MIT license) by Supermedium (formerly Mozilla VR) and other contributors.
\end{compactdesc}

\begin{figure}
\centering
\includegraphics[width=\columnwidth]{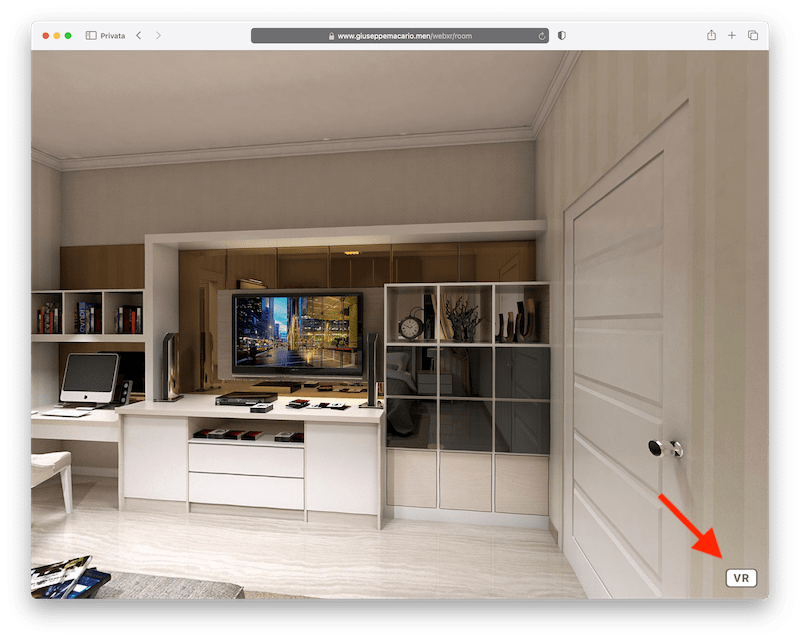}
\caption{VR button in the lower right corner of an A-Frame scene}
\label{fig:room}
\end{figure}

A constantly updated list of browsers supporting WebXR is available online\footnote{\url{https://caniuse.com/webxr}} and, at the time of writing (\the\year{}), it shows that the latest versions of Chrome, Edge, Opera, and Samsung Internet, fully support the API. Firefox and Safari also support the API---not yet on iOS\footnote{In the meantime, WebXR is supported by other iOS apps such as Mozilla's WebXR Viewer.}---although it may have to be manually enabled by the user. Consequently, all the most popular XR headsets also support WebXR, through their built-in browsers: Meta Quest, Microsoft HoloLens, Apple Vision Pro, HTC Vive, Samsung Gear, Google Cardboard and others. Also, desktop and mobile browsers support WebXR; however, since desktops, laptops, smartphones and smart TVs are not specifically designed for XR, the browser is forced to graphically simulate the immersive features of WebXR, depending on the screen and the platform. This is why A-Frame\footnote{See section~\ref{section:a-frame}.} apps can show a button in the lower right corner (fig.~\ref{fig:room})
that enables users to transition from a traditional 2D web environment into an immersive 3D experience.\footnote{In fig.~\ref{fig:room}, the button is labeled as ``VR'', but other implementations could use different labels.} This critical interface element, which provides users with a more engaging and interactive experience, can be displayed on any browser supporting WebXR. Depending on the needs of the developer, its behavior can be customized; in the case of desktop browsers, it usually toggles full screen mode.

\subsection{A-Frame}
\label{section:a-frame}

Three.js is an open-source WebGL-based cross-browser JavaScript library and API used for 3D graphics within web browsers. By abstracting the Three.js syntax into an HTML-like syntax through the entity-component-system (ECS) architectural pattern \citep{Wang2020CrossPlatformWebBasedVR}, the A-Frame framework makes WebGL\slash Three.js applications much easier and faster to code. The ECS is a popular and desirable pattern in 3D and game development, which follows the composition over inheritance and hierarchy principle \citep{Supermedium2024ECS}.

A-Frame provides the entire WebXR Device API, and is therefore suitable for XR devices. Through a web browser, the resulting applications run on common XR systems as well as conventional computers and mobile devices, thus reaching the largest possible number of users without the need for additional hardware or software.

In accordance with the ECS pattern, an A-Frame component is a modular and reusable data container that adds appearance and/or behavior to an entity. Therefore, an A-Frame-based virtual world is basically a combination of A-Frame entities and components, which can optionally load 3D models and multimedia assets such as images, sounds and videos.

\subsection{Networked-Aframe}

Since A-Frame is primarily geared towards single-user experiences, the Networked-Aframe framework builds upon A-Frame by introducing networked components that enable real-time, multi-user experiences within the same virtual space, mirroring the dynamics of physical users' interactions. Networked-Aframe communicates via WebRTC, a peer-to-peer API, or alternatively via WebSockets, a client-server API.

\section{Previous work}

According to \citet{Lombardi2010Opening}, the metaverse should
\begin{compactenum}
\item be entirely open and non-proprietary,
\item include a virtual-world browser similar to a web browser,
\item provide hyperlinks for traversing the virtual world.
\end{compactenum}
The authors implemented these recommendations in Open Cobalt, an open-source distributed platform designed for creating decentralized 3D virtual worlds and written in the Squeak programming language (a derivative of Smalltalk). However, Open Cobalt has not been actively maintained in recent years, and its usage has declined as other platforms and technologies for virtual reality and 3D interaction, like Unity or Unreal Engine, have become more popular and supported.

\citet{Junmyeong2022DevelopmentWebBased} observed that the accessibility of the metaverse is hindered by the requirement to install specific software or to use dedicated devices. Therefore, they developed a web-based metaverse that users can access simply by typing a URL into the address bar of their browser. The graphics and physics of their metaverse are powered by the Unity game engine, which is proprietary software.

\citet{Havele2022KeysOpen} argue that a truly comprehensive Metaverse should extend beyond isolated ``microverses'' or restrictive ``walled gardens''. This paradigm shift underscores a move from earlier, more fragmented virtual experiences to a more integrated and accessible virtual ecosystem. A ``Unified Metaverse'', they contend, would facilitate smooth transitions between different virtual environments without exiting a three-dimensional state, supported by robust interoperability standards. The authors propose that such a metaverse would naturally evolve from the current two-dimensional web, suggesting that it could emerge through enhanced web connectivity and the implementation of open web standards. They highlight X3D as a royalty-free ISO/IEC open standard for 3D computer graphics and interactive applications on the web, developed and maintained by the \citet{Web3DConsortium2024Standards}. It serves as a successor to the Virtual Reality Modeling Language (VRML) and provides a framework for representing a wide range of 3D graphics elements such as rendering, animation, and user interaction. X3D applets run within web browsers, using OpenGL technology to display 3D content across different browsers and operating systems. However, X3D has not achieved widespread acceptance, which may be due to the learning curve associated with its implementation and use.

\paragraph{Novel contribution} This paper introduces an approach that aligns with the foundational principles set forth by \citet{Lombardi2010Opening} for an open, non-proprietary virtual world, enhanced with a modern browser capable of navigating through interconnected spaces via hyperlinks. The proposed solution leverages current web technologies to ensure widespread compatibility and usability across most contemporary browsers and devices. By addressing the challenges of accessibility and interoperability highlighted in \cite{Junmyeong2022DevelopmentWebBased} and \cite{Havele2022KeysOpen}, this approach seeks to foster an open and integrated metaverse environment.

\section{A ``Hello world'' spatial web app}
\label{section:spatial}

The academic introduction of the term spatial computing is attributed to \citet{Greenwold2003SpatialComputing}:
\begin{definition}[Spatial computing]
Human interaction with a machine in which the machine retains and manipulates referents to real objects and spaces.
\end{definition}

According to \citet{Apple2023SpatialComputing}, ``Featuring visionOS, the world's first spatial operating system, Vision Pro lets users interact with digital content in a way that feels like it is physically present in their space'' and ``With more than 2,000 spatial apps designed for Apple Vision Pro [\dots\unkern] visionOS~2 enables developers to take further advantage of spatial computing'' \cite{Apple2024visionOS2}: in other words, spatial computing and spatial apps need a spatial operating system (visionOS, in this case).

On the other hand, in section~\ref{section:webxr}, we introduced the concept of ``spatial web apps''. By spatial web apps, we mean spatial apps that do not necessarily have to run on a spatial operating system such as visionOS. To run a spatial web app, users can simply open a URL in any browser; therefore, unlike Apple's spatial apps, spatial web apps can run everywhere.

Although no installation is required, a spatial web app can still be installed using features such as Progressive Web Apps (PWA) and ``Add to Home Screen'', which allows mobile websites to open like native apps, bypassing the Apple or Google App Stores. An app delivered as a regular web page will never need to be manually updated by the user because it is sufficient for the developer to update the web page. Moreover, the app automatically benefits from browser features such as history and tabs, which are already well known to the general public.

\begin{figure}
\centering
\includegraphics[width=.9\columnwidth]{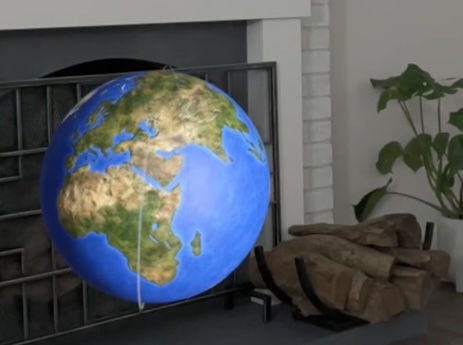}
\caption{``Hello world'' in visionOS and WebXR}
\label{fig:hello-world}
\end{figure}

\citet{Apple2024HelloWorld} provides an example of ``Hello world'' in visionOS, i.e. a spatial app that shows a floating globe (fig.~\ref{fig:hello-world}); this downloadable Xcode project contains 180 files, for a total of about 250~MB of source code.
However, Apple Vision Pro also supports WebXR through Safari for visionOS, which allows the developer to create a spatial web app without using Xcode and avoiding Apple's review process. Instead of compiling a project of 180 files, WebXR does the same thing in very few lines of A-Frame code---even just one line---within one HTML document:
\begin{lstlisting}[language=HTML,basicstyle=\footnotesize\sffamily,showstringspaces=false,columns=fullflexible,keywordstyle=]
<body>
 <a-scene>
 <a-sphere src="url(https://example.com/texture.jpg)"
        position="0 1.5 -5" radius="1" rotation="0 0 -30"
        animation="property: rotation; to: 0 360 -30;
                       loop: true; dur: 10000; easing: linear">
  </a-sphere>
 </a-scene>
</body>
\end{lstlisting}

In general, a spatial web app consists of at least an HTML document with JavaScript, possibly including textures and 3D models. A minimal valid HTML page with the necessary code is available online,\footnote{\url{https://www.giuseppemacario.men/webxr/hello-world}} where devices can also test the app.

Overall, this example demonstrates some of the advantages of a WebXR spatial web app over a native visionOS spatial app. Firstly, the app can run on any device, not just on Apple Vision Pro. Secondly, the source code of an HTML document is extremely light compared to a large Xcode project and does not need to be compiled. Furthermore, the spatial web app does not have to be distributed through an app store. These advantages also apply to other non-Apple headsets.
\section{Conceptual architecture}

\begin{figure}
\centering
\includegraphics[width=\columnwidth]{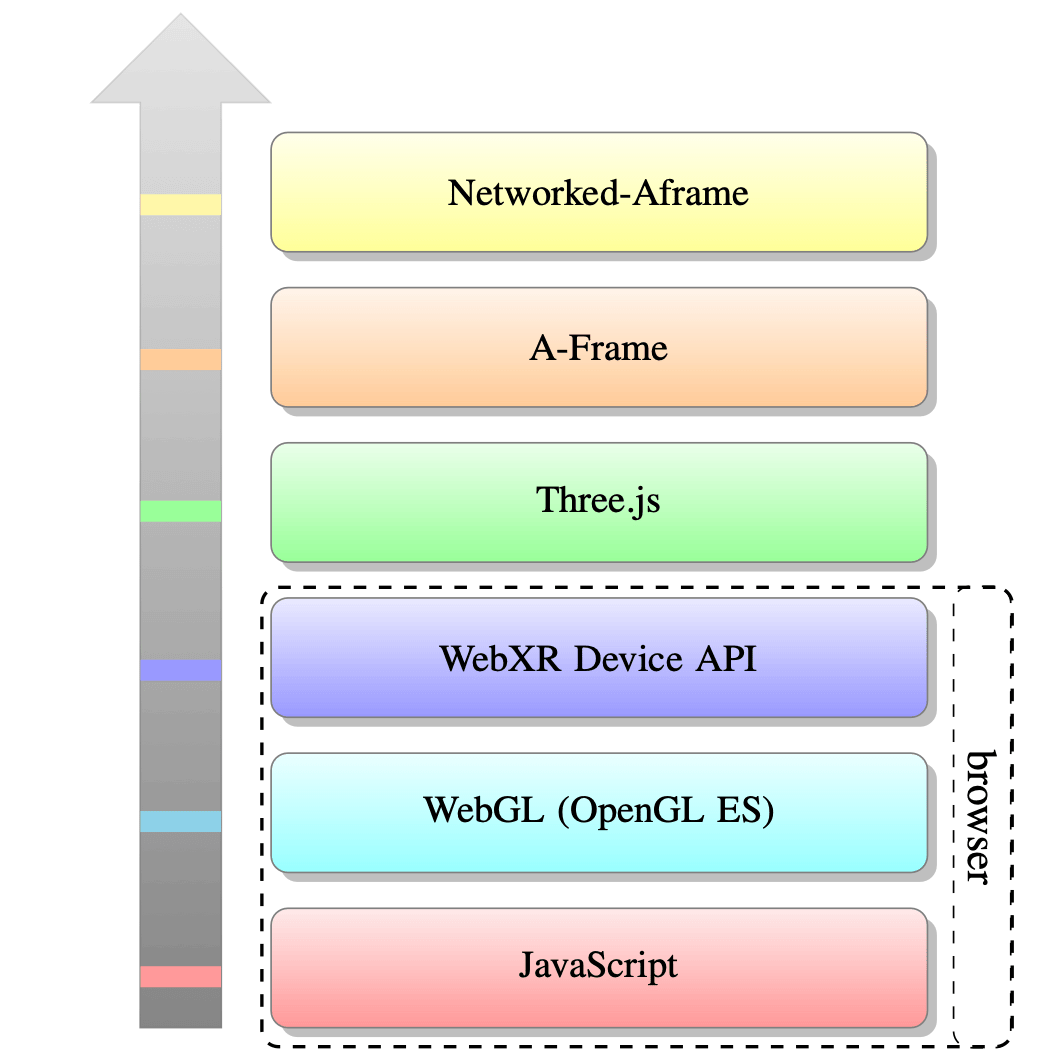}
\caption{Technology stack of a WebXR\slash A-frame spatial web app (abstraction layers in ascending order)}
\label{fig:webxr-stack}
\end{figure}

The conceptual architecture, as illustrated in fig.~\ref{fig:webxr-stack}, represents a layered approach to developing spatial web applications. Each layer builds upon the foundational technologies:
\begin{compactdesc}
\item[JavaScript] Serves as the primary programming language.
\item[WebGL (OpenGL ES)] Provides a low-level graphics API for rendering 2D and 3D graphics in the browser.
\item[WebXR Device API] Facilitates direct interaction with XR hardware.
\item[Three.js] Acts as a lightweight 3D library that builds on top of WebGL, simplifying 3D content creation.
\item[A-Frame] Provides a web framework for building VR experiences, making Three.js accessible to web developers without deep knowledge of WebGL.
\item[Networked-Aframe] Extends A-Frame for multiplayer experiences, enabling the development of a shared virtual environment.
\end{compactdesc}

As seen in section \ref{section:webxr}, the first three components are natively supported by web browsers; therefore, in fig.~\ref{fig:webxr-stack}, they are encapsulated within a ``browser'' environment, indicating the stack's capability to deliver immersive experiences directly through a web browser without the need for external applications.

\section{Comparison between Second Life and the proposed open metaverse}

Existing metaverse platforms can be represented by Second Life, which remains the oldest and most popular virtual world directly inspired by Neal Stephenson's concept of the Metaverse. It shares several technical features with the envisioned metaverse, such as customizable avatars \citep{SL2024Avatar,Stephenson1992SnowCrash} and regions sized as multiples of 256, specifically $256\,\text{m} \times 256\,\text{m} = 65{,}536\,\text{m}^2$ each \citep{SLWiki2024Region,Stephenson1992SnowCrash}. The pioneering role of Second Life in virtual worlds is unquestionable, but it exhibits several limitations when compared to the open metaverse proposed in this paper. The following points highlight several key areas where the new metaverse offers significant improvements; Second Life is used as a benchmark, but each point is applicable to other major virtual worlds as well.

\paragraph{Access to other virtual worlds}
Second Life confines users within its closed ecosystem, offering no direct interaction with other virtual worlds. In contrast, the proposed metaverse promises potentially unlimited access to a diverse array of virtual environments in a very simple way, namely through HTML links. To connect to a different virtual environment, one simply needs to click on a hyperlink, and the new environment will load in the current browser window or in a new window.\footnote{Depending on the presence of the attribute \textsf{target="\_blank"}.}

\paragraph{Supported platforms}
Second Life clients (also known as \emph{viewers}) run on Windows, macOS and Linux, while the proposed metaverse can be accessed with any web browser, significantly expanding the range of supported devices, including mobile ones.

\paragraph{In-world electronic currency}
Second Life operates with a proprietary in-world currency, which is limited to transactions within its platform. The new metaverse model, however, supports the use of various electronic currencies, including both fiat and cryptocurrencies, thereby significantly broadening the economic flexibility for users.

\paragraph{Protocol and server}
Second Life uses a proprietary protocol designed by Linden Lab; only a portion of it has been made public \citep{SLWiki2024Protocol}, although a reasonably compatible open-source protocol named libOpenMetaverse has been made available by reverse-engineering Linden Lab's protocol \citep{libOMV2024Github}. The primary server process in Second Life is called a \emph{simulator}, which simulates one region. Each CPU core is dedicated to one server process; therefore, a host\footnote{By metonymy, the term \emph{server} can also refer to a host.} can execute four or more server processes at once \citep{SLWiki2024Server} depending on the number of CPU cores it has. On the other hand, the proposed metaverse is designed to function on any web server using HTTPS, which simplifies deployment and reduces costs for developers who wish to host and manage their virtual spaces.

\paragraph{Costs for virtual land\slash space}
Owning virtual land in Second Life involves subscription fees set by Linden Lab, while the open architecture of the proposed metaverse allows any website to host a virtual space, enabling users to create and own virtual spaces either for free or at significantly reduced costs.

\paragraph{Encryption}
The security measures in Second Life are relatively limited, particularly concerning data encryption. The client encrypts user authentication (username and password) \citep{SLWiki2024Security} but everything else is transmitted over unencrypted UDP packets. On the other hand, the proposed metaverse uses HTTPS for all data exchanges. This standard practice for Internet safety greatly enhances the protection of user information and interactions, offering a more secure and trustworthy virtual environment.

\section{Evaluation}

\subsection{Method}

Performance tests were conducted using Cloudflare's infrastructure, which provides detailed logs and analytics to trace each request and response through the network. These tests were complemented by Google PageSpeed Insights\footnote{\url{https://pagespeed.web.dev}}. Static assets were distributed through Cloudflare’s Content Delivery Network (CDN), reducing load times and latency by caching assets at edge servers. This strategy minimizes latency by conducting computations geographically closer to the user, a critical factor for the immersive experiences offered by the virtual environment. To ensure the reproducibility of the tests, all experiments were performed using the default settings of Cloudflare's free plan, without any modifications. The complete source code of the spatial web apps and the deployment scripts are provided in the Italian Ministry of Defense's repository\footnote{\url{https://virtual.difesa.it}}, accessible through a personal account available for free upon request.

\subsection{User experience survey}

A user experience survey was conducted with 100 volunteer participants from the Multipurpose Research Center of the Ministry of Defense to gather qualitative feedback on the platform. The survey consisted of 10 questions designed to evaluate various aspects of the platform's usability and effectiveness. Responses were collected using a Likert scale from 1 to 5, with additional space for general feedback. Analyzing these responses helped refine the platform's design, focusing on user-driven improvements.

\subsubsection{Survey questions}
\begin{compactenum}
\item How would you rate the ease of navigation within the virtual environment? (1~Very difficult, 5~Very easy)
\item How would you rate the visual quality of the virtual environment? (1~Very poor, 5~Excellent)
\item How satisfied are you with the response time of the virtual environment? (1~Very unsatisfied, 5~Very satisfied)
\item How easy was it to access the virtual environment from different devices? (1~Very difficult, 5~Very easy)
\item How responsive did you find the multiplayer interactions? (1~Very unresponsive, 5~Very responsive)
\item How well do you think the virtual environment supports collaboration and interaction between users? (1~Not well, 5~Very well)
\item How secure did you feel using the virtual environment? (1~Not secure, 5~Very secure)
\item How satisfied are you with the overall user experience provided by the virtual environment? (1~Very unsatisfied, 5~Very satisfied)
\item How likely are you to recommend the virtual environment to others? (1~Not likely, 5~Highly likely)
\item Do you think this virtual environment is better or worse than proprietary virtual environments you already know? (1~Much worse, 5~Much better)
\item Optional: Please provide any additional feedback or suggestions.
\end{compactenum}

\subsection{Results}

\subsubsection{Performance metrics}

The following metrics provide a quantitative foundation for assessing the prototype’s performance, highlighting its strengths in handling complex scenes and real-time interactions, albeit with noted areas for improvement.

\paragraph{Load time}
Load times were measured across various devices, including five desktop computers (both high-performance and standard models) and five mobile devices (from high-end to mid-tier smartphones), from the initiation of the application to the full rendering of the 3D scene. The Largest Contentful Paint (LCP), a crucial user-centric metric for perceived loading speed, was evaluated using Google PageSpeed Insights. On average, the LCP consistently remained below 2.5 seconds for both desktop and mobile platforms, ranking above the 75th percentile. This performance level indicates superior efficiency in content delivery and optimization, well-aligned with current web standards.

\paragraph{Frame rate}
The frame rate was monitored using Chrome DevTools on a suite of five desktops and five mobile devices to ensure a smooth user experience across various user densities and scene complexities. The application consistently maintained a frame rate of over 60 FPS on both desktop and mobile platforms, often reaching up to 120 FPS. Despite this high performance, noticeable dips occurred when the user count exceeded 20, indicating areas where further optimization is necessary.

\paragraph{Network latency (ping)} Network latency was assessed to evaluate the responsiveness of multiplayer interactions. The delay revealed an average latency of 120 milliseconds, which is below the threshold that might disrupt real-time interactions. However, occasional spikes up to 300 milliseconds were observed during peak server load times.

\subsubsection{Survey results}

\begin{compactdesc}
\item[Ease of navigation] 85\% of participants rated navigation as easy or very easy.
\item[Visual quality] 88\% rated the visual quality as high or very high.
\item[Response time] 79\% are satisfied or very satisfied with the response time.
\item[Cross-device accessibility] 91\% consider that accessing the platform from different devices was easy or very easy.
\item[Multiplayer responsiveness] 82\% found the multiplayer interactions to be responsive or very responsive.
\item[Collaboration and interaction between users] 77\% found the virtual environment to be collaborative.
\item[Security] 83\% found the virtual environment to be secure or very secure.
\item[Overall satisfaction] 87\% are satisfied or very satisfied with the user experience.
\item[Recommendation] 85\% are likely or very likely to recommend the platform to others.
\item[Comparison with proprietary virtual worlds] 92\% found this virtual environment to be better or much better than proprietary virtual environments they know.
\item[Additional feedback] Common suggestions included enhancing avatar customization options and improving performance on smart TVs.
\end{compactdesc}

\section{Discussion}

The evaluation showed that the prototype is robust, supporting immersive virtual spaces across multiple devices while maintaining a high quality of user experience. This underscores the practical viability of the conceptual architecture and highlights its potential as a foundational framework for developing open metaverse environments. Technical performance was generally strong, though there is room for improvement in handling complex scenes with high user density. Users gave positive feedback about the ease of use and the immersive quality of the environment.

While the prototype showcased the architecture's capabilities, several limitations emerged. Insights into cross-platform accessibility indicate a need for further refinement to optimize the experience on slower devices (smart TVs). Performance issues under high-density conditions highlighted the need for optimization, particularly in terms of resource management and scene complexity. Network latency, although acceptable, remains a concern for real-time interactions, suggesting that further improvements in synchronization and data transmission are necessary.

The insights gained from this research suggest various directions for future work:

\paragraph{Optimization} Dynamic content loading and rendering optimizations could enhance performance, especially on constrained hardware. These strategies may involve more efficient use of resources and improved algorithms for real-time processing.

\paragraph{User interface (UI) and User experience (UX) design} Exploring innovative UI/UX designs could further improve usability and accessibility, making virtual spaces more intuitive and engaging for a broader audience. This could include the integration of adaptive interfaces that adjust to user preferences and needs.

\paragraph{Multiuser interactions} Investigating advanced techniques for networked interactions and data synchronization could enhance the realism and fluidity of multiplayer experiences. This might involve the development of new protocols or the refinement of existing ones to reduce latency and improve the consistency of shared virtual spaces.

\paragraph{Scalability} Examining architectural modifications or extensions to support larger, more complex environments without compromising performance would be valuable. Potential approaches could include the use of distributed computing architectures or cloud-based rendering solutions.

\paragraph{Future features of WebXR} Looking ahead, the WebXR Device API is going to expand its functionality to include advanced features that will further enrich immersive web experiences. These anticipated enhancements include the implementation of world anchors, which would allow digital content to be consistently positioned within the physical world, and hit-testing capabilities, enabling interaction between virtual objects and the real-world environment detected by platform sensors. Additionally, the API is expected to facilitate the exposure of world structure, offering developers a deeper understanding of the physical space around the user through data from platform sensors.

\section{Conclusion}

This research aimed to address the persistent fragmentation in the metaverse by proposing a WebXR-based cross-platform architecture using the A-Frame and Networked-Aframe frameworks. The development and evaluation of a prototype demonstrated this architecture's ability to facilitate engaging and interactive virtual spaces, available across various platforms and devices, reflecting a significant step toward an open, interoperable metaverse. The architecture's reliance on open web standards supports Tim Berners-Lee's vision for an accessible and universal web, and ensures that the metaverse remains an inclusive digital ecosystem, free from the constraints of proprietary technologies.

The prototype was thoroughly evaluated, receiving positive feedback particularly for its ease of use and the immersive quality of the environment. This feedback underscores the practical viability of the proposed solution for creating a user-friendly metaverse. Performance tests further validated the robustness of the architecture, with data from Cloudflare’s infrastructure and Google PageSpeed Insights contributing to a better understanding of network efficiency and responsiveness. The application maintained excellent load times and frame rates, important metrics for immersive experiences. Additionally, the implementation of the HTTPS protocol for security aligns with best practices for protecting user data, reinforcing the architecture’s suitability for widespread adoption.

Looking forward, further research will be essential to explore advanced features of WebXR and other emerging web technologies that could enrich the metaverse experience. Implementing such technologies could potentially expand the scope of interactive and collaborative possibilities within the metaverse, which moves us closer to a truly interconnected digital world that interfaces with the physical world through extended reality.

%\bigskip
%{
%\small
%\textbf{Declaration of generative AI and AI-assisted technologies in the writing process}
%
%During the preparation of this work the author(s) used ChatGPT in order to improve the translation and the readability of the paper. After using this tool/service, the author(s) reviewed and edited the content as needed and take(s) full responsibility for the content of the published article.
%}

\bibliographystyle{ACM-Reference-Format}
\bibliography{mybibl}
\end{document}